\documentclass{llncs}

\usepackage{latexsym,amssymb,amsmath}
\usepackage{algorithm}
\usepackage[noend]{algpseudocode}
\usepackage{xspace}
\usepackage{url}
\usepackage{enumerate}
\usepackage[defblank]{paralist}
\usepackage{graphicx}
\graphicspath{{imgs/}}
\usepackage{todonotes}
\usepackage[defblank]{paralist}
 \usepackage{comment}
 \usepackage{tabularx}
 \usepackage{thesisSymbols}
 \usepackage{booktabs}
 \usepackage{multirow}
\allowdisplaybreaks[2]

\usepackage{times}

\newcommand{\myi}{(\emph{i})\xspace}
\newcommand{\myii}{(\emph{ii})\xspace}

\newcommand{\A}{\mathcal{A}} 
\newcommand{\C}{\mathcal{C}}

 \renewcommand{\L}{\mathcal{L}}
\newcommand{\M}{\mathcal{M}} 
 \renewcommand{\P}{\mathcal{P}}
 
\renewcommand{\S}{\mathcal{S}} 
\newcommand{\U}{\mathcal{U}}

\newcommand{\limp}{\mathbin{\rightarrow}}

\newcommand{\Next}{\raisebox{-0.27ex}{\LARGE$\circ$}}
\newcommand{\Wnext}{\raisebox{-0.27ex}{\LARGE$\bullet$}}
\renewcommand{\Until}{\mathop{\U}}

\newcommand{\true}{\mathit{true}}

\newcommand{\false}{\mathit{false}}
\newcommand{\ttrue}{\mathtt{true}}
\newcommand{\ffalse}{\mathtt{false}}
\newcommand{\Last}{\mathit{last}}

\newcommand{\last}{\mathit{n}}
\newcommand{\nnf}{\mathit{nnf}}

\newcommand{\BOX}[1]{ [#1]}
\newcommand{\DIAM}[1]{\langle #1 \rangle}

\newcommand{\LTL}{{\sc ltl}\xspace}
\newcommand{\LTLf}{{\sc ltl}$_f$\xspace}

\newcommand{\LDL}{{\sc ldl}\xspace}
\newcommand{\LDLf}{{\sc ldl}$_f$\xspace}
\newcommand{\RE}{{\sc re}$_f$\xspace}
\newcommand{\PDL}{{\sc pdl}\xspace}

\newcommand{\AFW}{{\sc afw}\xspace}
\newcommand{\NFA}{{\sc nfa}\xspace}
\newcommand{\DFA}{{\sc dfa}\xspace}

\newcommand{\declare}{{\sc declare}\xspace}

\newcommand{\re}{\mathit{re}}

\newcommand{\Nat}{{\rm I\kern-.23em N}}
\newcommand{\Prop}{\P}

\newcommand{\pref}{\mathsf{pref}}
\newcommand{\temptrue}{\mathit{temp\_true}}
\newcommand{\tempfalse}{\mathit{temp\_false}}
\newcommand{\possgood}{\mathit{poss\_good}}

\newcommand{\necgood}{\mathit{nec\_good}}
\newcommand{\necbad}{\mathit{nec\_bad}}

\renewcommand{\ttrue}{\mathit{tt}}
\renewcommand{\ffalse}{\mathit{ff}}
\newcommand{\Endt}{\mathit{end}}
\newcommand{\atomize}[1]{\texttt{"}\ensuremath{#1}\texttt{"}}

\newcommand{\ot}{o}

\newenvironment{proofsk}{\noindent\textsl{Proof (sketch).\ }}{\qed}

\makeatletter
\newcommand*{\inlineequation}[2][]{%
  \begingroup
    \refstepcounter{equation}%
    \ifx\\#1\\%
    \else
      \label{#1}%
    \fi
    \relpenalty=10000 %
    \binoppenalty=10000 %
    \ensuremath{%
      #2%
    }%
    ~\@eqnnum
  \endgroup
}
\makeatother

\newif\ifdraft
\drafttrue

\title{LTLf and LDLf Monitoring}
\subtitle{A Technical Report}

\author{
 Giuseppe De Giacomo\inst{1} \and Riccardo De Masellis\inst{1} \and Marco Grasso\inst{1}
 \and \\ Fabrizio Maria Maggi\inst{2} \and Marco Montali\inst{3}
}

\institute{
Sapienza Universit\`a\  di Roma,
Via Ariosto, 25, 00185 Rome, Italy\\
\email{(degiacomo|demasellis)@dis.uniroma1.it}
\and
University of Tartu,
J. Liivi 2, 50409 Tartu, Estonia\\
  \email{f.m.maggi@ut.ee}
\and
Free University of Bozen-Bolzano,
Piazza Domenicani 3, 39100 Bolzano, Italy\\
\email{montali@inf.unibz.it}
}

\begin{document}

\maketitle
\sloppy

\begin{abstract}
Runtime monitoring is one of the central tasks to provide operational
decision support to running business processes, and check on-the-fly
whether they comply with constraints and rules.
We study runtime monitoring of properties expressed in \LTL on finite traces (\LTLf) and in its extension \LDLf.
\LDLf is a powerful logic that captures all monadic second order logic on finite traces, which is obtained by combining regular expressions and \LTLf, adopting the syntax of propositional dynamic logic (\PDL). Interestingly, in spite of its greater expressivity, \LDLf has exactly the same computational complexity of \LTLf.
We show that \LDLf is able to capture, in the logic itself,
not only the constraints to be monitored, but also the de-facto standard RV-LTL monitors. This makes it possible to declaratively capture monitoring
metaconstraints, and check them by relying on usual logical services
instead of ad-hoc algorithms. This, in turn, enables to flexibly
monitor constraints depending on the monitoring state of
other constraints, e.g., ``compensation'' constraints that are only
checked when others are detected to be violated.
In addition, we devise a direct translation of \LDLf formulas into
nondeterministic automata, avoiding to detour to B\"uchi automata or alternating automata, and we use it to implement a monitoring plug-in for the \textsc{ProM} suite.
\end{abstract}


\section{Introduction}
Runtime monitoring is one of the central tasks to provide \emph{operational
decision support} \cite{Aal11} to running business processes, and check on-the-fly
whether they comply with constraints and rules. In order to provide
well-founded and provably correct runtime monitoring techniques, this
area is usually rooted into that of \emph{verification}, the branch of formal analysis
aiming at checking whether a system meets some property of
interest. Being the system dynamic, properties are usually
expressed by making use of modal operators accounting for the time.



Among all the temporal logics used in verification, Linear-time
Temporal Logic (\LTL) is particularly suited for monitoring, as an
actual system execution is indeed linear. However, the 
\LTL semantics is given in terms of infinite traces, hence monitoring must check whether the current trace is a prefix of an infinite trace, that will never be completed \cite{Bauer2010:LTL}.
In several context, and in particular often in BPM, we can assume that
the trace of the system is if face finite \cite{PesV06}. For this reason, finite-trace variant of the \LTL have been introduced. 
Here we use the logic \LTLf (LTL on finite traces), investigated in
detail in \cite{DegVa13}, and at the base of one of the main
declarative process modeling approaches: \declare \cite{PesV06,MPVC10,MMW11}.
Following \cite{MMW11}, monitoring in \LTLf amounts to
checking whether the current execution belongs to the set of
admissible \emph{prefixes} for the traces of a given \LTLf formula
$\varphi$. To achieve such a task, $\varphi$ is usually first translated
into a finite-state automaton for $\varphi$, which recognizes all those
\emph{finite} executions that satisfy $\varphi$.

Despite the presence of previous operational decision support techniques to monitoring \LTLf
constraints over finite traces \cite{MMW11,MWM12}, two main
challenges have not yet been tackled in a systematic way. First of
all, several alternative semantics have been proposed to make LTL
suitable for runtime verification (such as the de-facto standard RV
monitor conditions \cite{Bauer2010:LTL}), but no
comprehensive technique based on finite-state automata is available to
accommodate them. On the one hand, runtime verification for such
logics typically considers finite partial traces whose continuation is
however infinite \cite{Bauer2010:LTL}, with the consequence that the corresponding
techniques detour to B\"uchi automata for building the monitors. On
the other hand, the incorporation of such semantics in the BPM setting
(where also continuations are finite) has only been tackled so far
with effective but ad-hoc techniques (cf.~the ``coloring'' of automata
in \cite{MMW11} to support the RV conditions), without a
corresponding formal underpinning. 

A second, key challenge is the incorporation of advanced forms of
monitoring, where some constraints become of interest only
in specific, critical circumstances (such as the violation of other
constraints). This is the basis for supporting monitoring of
compensation constraints and so-called contrary-to-duty obligations
\cite{PrS96}, i.e., obligations that are put in place only when other
obligations have not been fulfilled. While this feature is considered
to be a fundamental compliance monitoring functionality \cite{LMM13}, it is
still an open challenge, without any systematic approach able to
support it at the level of the constraint specification language. 

In this paper, we attack these two challenges by studying runtime monitoring of properties expressed in \LTLf  and in its extension \LDLf \cite{DegVa13}.
\LDLf is a powerful logic that captures all monadic second order logic on finite traces, which is obtained by combining regular expressions and \LTLf, adopting the syntax of propositional dynamic logic (\PDL). Interestingly, in spite of its greater expressivity, \LDLf has exactly the same computational complexity of \LTLf.
We show that \LDLf is able to capture, in the logic itself,
not only the usual \LDLf constraints to be monitored,
 but also the de-facto standard RV monitor conditions. 
Indeed given an \LDLf formula $\phi$, we show how to construct
the \LDLf formulas that captures whether prefixes of $\phi$ satisfy the various RV monitor conditions.
This, in turn, makes it possible to declaratively capture \emph{monitoring
metaconstraints}, and check them by relying on usual logical services
instead of ad-hoc algorithms. Metaconstraints provide a well-founded, declarative basis
to specify and monitor constraints depending on the monitoring state of
other constraints, such as ``compensation'' constraints that are only
checked when others are 
violated.

Interestingly, in doing so we devise a direct translation of \LDLf (and hence of \LTLf)
formulas into nondeterministic automata, which avoid the usual detour
to B\"uchi automata. The technique is grounded on
alternating automata (\AFW), but it actually avoids also their introduction all
together, and directly produces a standard non-deterministic
finite-state automaton (\NFA).
Notably, such technique has been concretely implemented, and embedded
into a monitoring plug-in for the \textsc{ProM}, which supports checking 
\LDLf constraints and metaconstraints. 



\section{\LTLf and \LDLf}
\label{sec:LTLf-LDLf}

In this paper we will adopt the standard \LTL and its variant \LDL interpreted over on finite runs.

\LTL on finite traces, called \LTLf \cite{DegVa13}, has exactly the same syntax as \LTL on infinite traces \cite{Pnueli77}. Namely, given a set of $\Prop$ of propositional symbols, \LTLf formulas are obtained through the following:
\[\varphi ::= \phi \mid \lnot \varphi \mid \varphi_1\land \varphi_2
\mid \varphi_1\lor \varphi_2  \mid \Next\varphi \mid \Wnext\varphi
\mid \Diamond\varphi \mid \Box\varphi \mid \varphi_1\Until\varphi_2\]
where $\phi$ is a propositional formuala over $\Prop$, $\Next$ is the \emph{next} operator, $\Wnext$ is  \emph{weak next}, 
$\Diamond$ is  \emph{eventually}, $\Box$ is  \emph{always}, $\Until$ is  \emph{until}.

It is known that \LTLf is as expressive as First Order Logic over finite traces, so strictly less expressive than regular expressons which in turn are as expressive as Monadic Second Order logic over finite traces.  On the other hand,  regular expressions are a too low
level formalism for expressing temporal specifications, since, for example, they miss a direct construct for negation and for conjunction~\cite{DegVa13}.

To overcome this difficulties, in \cite{DegVa13}
\emph{Linear Dynamic Logic of Finite Traces}, or \LDLf, has been proposed. This logic
is as natural as \LTLf but with the full expressive power of Monadic Second Order logic over finite traces.  \LDLf is obtained by merging \LTLf with regular expression through the 
syntax of the well-know logic of programs \PDL, \emph{Propositional Dynamic
Logic}, 
\cite{FiLa79,HaKT00} but adopting a semantics based
on finite traces.  This logic  is an adaptation of \LDL, introduced in 
\cite{Var11}, which, like \LTL, is interpreted over infinite traces.

Formally, \LDLf formulas are built as follows:
\[\begin{array}{lcl}
\varphi &::=& \phi \mid \ttrue \mid \ffalse\mid \lnot \varphi \mid \varphi_1 \land \varphi_2 \mid \varphi_1 \land \varphi_2 \mid \DIAM{\rho}\varphi \mid \BOX{\rho}\varphi\\
\rho &::=& \phi \mid \varphi? \mid  \rho_1 + \rho_2 \mid \rho_1; \rho_2 \mid \rho^*
\end{array}
\]
where $\phi$ is a propositional formula over $\Prop$; $\ttrue$ and
$\ffalse$ denote respectively the true and the false \LDLf formula
(not to be confused with the propositional formula $\true$ and
$\false$), $\rho$ denotes path expressions, which are \RE expressions
over propositional formulas $\phi$, with the addition of the test
construct $\varphi?$ typical of \PDL; and $\varphi$ stand for \LDLf
formulas built by applying boolean connectives and the modal
connectives $\DIAM{\rho}\varphi$ and $\BOX{\rho}\varphi$.  In fact
$\BOX{\rho}\varphi\equiv\lnot\DIAM{\rho}{\lnot\varphi}$.

Intuitively, $\DIAM{\rho}\varphi$ states that, from the current step in the trace,
there exists an execution satisfying the regular expression $\rho$
such that its last step satisfies $\varphi$. While
$\BOX{\rho}\varphi$ states that, from the current step, all
executions satisfying the regular expression $\rho$ are such that
their last step satisfies $\varphi$.
Tests are used to insert into the execution path checks for
satisfaction of additional \LDLf formulas.

As for \LTLf, the semantics of \LDLf is given in terms of \emph{finite traces} denoting a
finite, possibly empty, sequence of consecutive steps in the trace, i.e., finite words
$\pi$ over the alphabet of $2^\Prop$, containing all possible
propositional interpretations of the propositional symbols in $\Prop$.
We denote by $\pi(i)$ the $i$th step in the trace. If the trace is shorter and  does not include an $i$th step  $\pi(i)$ is undefined.
We denote by $\pi(i,j)$ the segment of the trace $\pi$ starting at $i$th step end ending at the $j$th step (included). If $i$ or $j$ are out of range wrt the trace that $\pi(i,j)$ is undefined, except $\pi(i,i)=\epsilon$ (i.e., the empty trace). 
 
The semantics of \LDLf is as follows:  an \LDLf formula $\varphi$ \emph{is true} 
at a step $i$,  in symbols $\pi,i\models\varphi$, as follows:
\begin{compactitem} 
\item $\pi,i\models \ttrue$
\item $\pi,i\not\models \ffalse$
\item $\pi,i\models \phi$ ~iff~ $1\leq i\leq \last$ and $\pi(i)\models \phi$ \quad ($\phi$ propositional).
\item $\pi,i\models \lnot\varphi$ ~iff~ $\pi,i\not\models\varphi$.
\item $\pi,i\models \varphi_1\land\varphi_2$ ~iff~ $\pi,i\models\varphi_1$ and
  $\pi,i\models\varphi_2$.
\item $\pi,i\models \varphi_1\lor\varphi_2$ ~iff~ $\pi,i\models\varphi_1$ or
  $\pi,i\models\varphi_2$.
\item $\pi,i\models \DIAM{\rho}\varphi$ ~iff~  for some $i$
we have $\pi(i,j)\in \L(\rho)$ and $\pi,j\models\varphi$.
\item $\pi,i\models \BOX{\rho}\varphi$ ~iff~  for all $i$ 
such that $\pi(i,j)\in \L(\rho)$ we have $\pi,j\models\varphi$.
\end{compactitem}
The relation $\pi(i,j)\in\L(\rho)$ 
 is defined inductively as follows:
\begin{compactitem}
\item $\pi(i,j)\in\L(\phi)$ if $ j=i+1 \leq \last
                                         \;\text{and}\; \pi(i)\models \phi  \quad \mbox{($\phi$ propositional)}$
\item $\pi(i,j)\in\L(\varphi?)$ if $j=i\;\text{and}\;  \pi, i\models \varphi$
\item $\pi(i,j)\in\L(\rho_1+ \rho_2)$ if $\pi(i,j)\in\L(\rho_1)\;\text{or}\; \pi(i,j)\in\L(\rho_2)$
\item $\pi(i,j)\in\L(\rho_1; \rho_2)$ if  $\mbox{ exists } k \mbox{  s.t.\  } \pi(i,k)\in\L(\rho_1) \mbox{ and } \pi(k,j)\in\L(\rho_2)$
\item
$\pi(i,j)\in\L(\rho^*)$ if $j=i\; \text{or} \mbox{ exists } k \mbox{ s.t.\  } \pi(i,k)\in\L(\rho) \mbox{ and } \pi(k,j)\in\L(\rho^*)$
\end{compactitem}

Observe that for $i>\last$, hence e.g., for $\pi=\epsilon$ we get:
\begin{compactitem} 
\item $\pi,i\models \ttrue$
\item $\pi,i\not\models \ffalse$
\item $\pi,i\lnot\models \phi$ \quad ($\phi$ propositional).
\item $\pi,i\models \lnot\varphi$ ~iff~ $\pi,i\not\models\varphi$.
\item $\pi,i\models \varphi_1\land\varphi_2$ ~iff~ $\pi,i\models\varphi_1$ and $\pi,i\models\varphi_2$.
\item $\pi,i\models \varphi_1\lor\varphi_2$ ~iff~ $\pi,i\models\varphi_1$ or  $\pi,i\models\varphi_2$.
\item $\pi,i\models \DIAM{\rho}\varphi$ ~iff~  $\pi(i,i)\in \L(\rho)$ and $\pi,i\models\varphi$.
\item $\pi,i\models \BOX{\rho}\varphi$ ~iff~  $\pi(i,i)\in \L(\rho)$ implies $\pi,i\models\varphi$.
\end{compactitem}
The relation $\pi(i,i)\in\L(\rho)$ with $i > \last$
 is defined inductively as follows:
\begin{compactitem}
\item $\pi(i,i)\not\in\L(\phi) \quad \mbox{($\phi$ propositional)}$
\item $\pi(i,i)\in\L(\varphi?)$ if $\pi, i\models \varphi$
\item $\pi(i,i)\in\L(\rho_1+ \rho_2)$ if $\pi(i,i)\in\L(\rho_1)\;\text{or}\; \pi(i,i)\in\L(\rho_2)$
\item $\pi(i,i)\in\L(\rho_1; \rho_2)$ if  $\pi(i,i)\in\L(\rho_1) \mbox{ and } \pi(i,i)\in\L(\rho_2)$
\item $\pi(i,i)\in\L(\rho^*)$
\end{compactitem}

Notice we have the usual boolean equivalences such as $\varphi_1\lor\varphi_2 \equiv \lnot \varphi_1\land \lnot \varphi_2$, furthermore we have that:
$\phi\equiv\DIAM{\phi}\ttrue$, and 
$\BOX{\rho}\varphi \equiv \lnot\DIAM{\rho}\lnot\varphi$.
It is also convenient to introduce the following abbreviations:
\begin{compactitem}
\item $\Endt=\BOX{\true?}\ffalse$ that denotes that the traces is been completed (the remaining trace is $\epsilon$ the empty one) 
\item $\Last= \DIAM{\true}\Endt$, which denotes the last step of the trace.
\end{compactitem}

It easy to encode \LTLf into  \LDLf: it suffice to observe that we can express the various \LTLf operators by recursively applying the following translations:

\begin{compactitem}
\item $\Next \varphi$ translates to $\DIAM{\true} \varphi$;
\item $\Wnext \varphi$ translates to $\lnot \DIAM{\true} \lnot\varphi=\BOX{true}\varphi$ (notice that $\Wnext a$ is translated into $\BOX{\true} \BOX{\lnot a}\ffalse$, since $a$ is equivalent to $\DIAM{a}\ttrue$);
\item $\Diamond \varphi$ translates to $\DIAM{\true^*} \varphi$;
\item $\Box \varphi$ translates to $\BOX{\true^*}\varphi$  (notice that $\Box a$ is translated into $\BOX{\true^*} \BOX{\lnot a}\ffalse$);
\item $\varphi_1  \Until  \varphi_2$ translates to $\DIAM{(\varphi_1?;\true)^*} \varphi_2$.
\end{compactitem}

It is also easy to encode regular expressions, used as a specification formalism for traces into \LDLf:
$\rho$ translates to $\DIAM{\rho} \Endt$.

We say that a trace satisfies an \LTLf or \LDLf formula $\varphi$, written $\pi\models \varphi$ if $\pi,1\models \varphi$.
(Note that if $\pi$ is the empty trace, and hence $1$ is out of range, still the notion of $\pi,1\models \varphi$ is well defined).
Also sometimes we denote  by $\L(\varphi)$ the set of  traces that satisfy $\varphi$:
$\L(\varphi) = \{\pi\mid \pi \models \varphi\}$.


\section{\LDLf Automaton}
\label{sec:automaton}
We can associate with each \LDLf formula $\varphi$ an (exponential)
\NFA $A_\varphi$ that accepts exactly the traces that make $\varphi$
true.
Here, we provide a simple direct algorithm for computing the \NFA
corresponding to an \LDLf formula.  The correctness of the algorithm
is based on the fact that \myi we can associate with each \LDLf
formula $\varphi$ a polynomial \emph{alternating automaton on words}
(\AFW) $\A_\varphi$ which accepts exactly the traces that make $\varphi$
true \cite{DegVa13}, and \myii every \AFW can be transformed into an
\NFA, see, e.g., \cite{DegVa13}. 
However, to formulate the algorithm we do not need these notions, but
we can work directly on the \LDLf formula.
In order to proceed with the construction of the \AFW $\A_\varphi$, we
put \LDLf formulas $\varphi$ in negation normal form $\nnf(\varphi)$
by exploiting equivalences and pushing negation inside as much as
possible, until is eliminated except in propositional formulas. Note
that computing $\nnf(\varphi)$ can be done in linear time. In other
words, wlog, we consider as syntax for \LDLf the one in the previous
section but without negation.
Then we define an auxiliary function $\delta$ that takes an \LDLf
formula $\psi$ (in negation normal form) and a propositional
interpretation $\Pi$ for $\Prop$ (including $\Last$), or a special symbol $\epsilon$, returning a
positive boolean formula whose atoms are (quoted) $\psi$ subformulas.
\begin{align*}
\delta(\atomize{\ttrue},\Pi) & = \true\\
\delta(\atomize{\ffalse},\Pi) & = \false\\
\delta(\atomize{\phi},\Pi) & = 
\left\{\begin{array}{l}
\true \mbox{ if } \Pi\models \phi \\
 \false \mbox{ if } \Pi\not\models \phi
\end{array}
\right.\quad \mbox{($\phi$ propositional)}\\
\delta(\atomize{\varphi_1\land\varphi_2},\Pi) & = 
\delta(\atomize{\varphi_1},\Pi) \land \delta(\atomize{\varphi_2},\Pi)\\
\delta(\atomize{\varphi_1\lor\varphi_2},\Pi) & =
\delta(\atomize{\varphi_1},\Pi) \lor \delta(\atomize{\varphi_2},\Pi)\\
\delta(\atomize{\DIAM{\phi}\varphi},\Pi) & = 
\left\{\hspace{-1ex}\begin{array}{l}
\atomize{\varphi} \mbox{ if } \Last \not \in \Pi \mbox{ and } \Pi \models \phi \quad \mbox{($\phi$  propositional)}\\
\delta(\atomize{\varphi}, \epsilon)  \mbox{ if } \Last \in \Pi \mbox{ and } \Pi \models \phi \\
\false \mbox{ if } \Pi\not\models \phi
\end{array}\right.\\
\delta(\atomize{\DIAM{\psi?}{\varphi}},\Pi) & = 
\delta(\atomize{\psi},\Pi) \land \delta(\atomize{\varphi},\Pi)\\
\delta(\atomize{\DIAM{\rho_1+\rho_2}{\varphi}},\Pi) & = 
\delta(\atomize{\DIAM{\rho_1}\varphi},\Pi) \lor \delta(\atomize{\DIAM{\rho_2}\varphi},\Pi)\\
\delta(\atomize{\DIAM{\rho_1;\rho_2}{\varphi}},\Pi) & = 
\delta(\atomize{\DIAM{\rho_1}\DIAM{\rho_2}\varphi},\Pi)\\
\delta(\atomize{\DIAM{\rho^*}\varphi},\Pi) & =
\left\{\hspace{-1ex}\begin{array}{l}
\delta(\atomize{\varphi},\Pi) \hfill \mbox{if  $\rho$ is test-only}\\
\delta(\atomize{\varphi},\Pi) \lor \delta(\atomize{\DIAM{\rho}\DIAM{\rho^*}\varphi},\Pi)
\;\;\hfill  \mbox{o/w}
\end{array}\right.\\
\delta(\atomize{\BOX{\phi}\varphi},\Pi) & =
\left\{\hspace{-1ex}\begin{array}{l}
    \atomize{\varphi} \mbox{ if } \Last \not \in \Pi \mbox{ and } \Pi \models \phi \quad \mbox{($\phi$ propositional)}\\
 \delta(\atomize{\varphi},\epsilon) \mbox{ if } \Last \in \Pi \mbox{ and } \Pi \models \phi \quad \mbox{($\phi$ propositional)}\\
    \true \mbox{ if } \Pi\not\models \phi
\end{array}\right.\\
\delta(\atomize{\BOX{\psi?}{\varphi}},\Pi) & = 
\delta(\atomize{\nnf(\lnot\psi)},\Pi) \lor \delta(\atomize{\varphi},\Pi)\\
\delta(\atomize{\BOX{\rho_1+\rho_2}{\varphi}},\Pi) & =  
\delta(\atomize{\BOX{\rho_1}\varphi},\Pi)\land\delta(\atomize{\BOX{\rho_2}\varphi},\Pi)\\
\delta(\atomize{\BOX{\rho_1;\rho_2}{\varphi}},\Pi) & = 
\delta(\atomize{\BOX{\rho_1}\BOX{\rho_2}\varphi},\Pi)\\
\delta(\atomize{\BOX{\rho^*}\varphi},\Pi) & = 
\left\{\hspace{-1ex}
\begin{array}{ll}
\delta(\atomize{\varphi},\Pi) \hfill \mbox{if  $\rho$ is test-only}\\
\delta(\atomize{\varphi},\Pi) \land \delta(\atomize{\BOX{\rho}\BOX{\rho^*}\varphi},\Pi) 
\;\;\hfill \mbox{o/w}
\end{array}\right.
\end{align*}
where $\delta(\atomize{\varphi},\epsilon)$, i.e., the interpretation of \LDLf formula in the case the (remaining fragment of the) trace is empty, is defined as follows:
\begin{align*}
\delta(\atomize{\ttrue},\epsilon) &= \true\\
\delta(\atomize{\ffalse},\epsilon) &= \false\\
\delta(\atomize{\phi},\epsilon) &= \false  \quad \mbox{($\phi$
  propositional)}\\
\delta(\atomize{\varphi_1\land\varphi_2},\epsilon) &= 
\delta(\atomize{\varphi_1},\epsilon) \land \delta(\atomize{\varphi_2},\epsilon)\\
\delta(\atomize{\varphi_1\lor\varphi_2},\epsilon) &= 
\delta(\atomize{\varphi_1},\epsilon) \lor \delta(\atomize{\varphi_2},\epsilon)\\
\delta(\atomize{\DIAM{\phi}\varphi},\epsilon) &=\false \quad \mbox{($\phi$  propositional)}\\
\delta(\atomize{\DIAM{\psi?}{\varphi}},\epsilon) &= \delta(\atomize{\psi},\epsilon) \land \delta(\atomize{\varphi},\epsilon)\\
\delta(\atomize{\DIAM{\rho_1+\rho_2}{\varphi}},\epsilon) &= 
\delta(\atomize{\DIAM{\rho_1}\varphi},\epsilon) \lor \delta(\atomize{\DIAM{\rho_2}\varphi},\epsilon)\\
\delta(\atomize{\DIAM{\rho_1;\rho_2}{\varphi}},\epsilon) &= 
\delta(\atomize{\DIAM{\rho_1}\DIAM{\rho_2}\varphi},\epsilon)\\
\delta(\atomize{\DIAM{\rho^*}\varphi},\epsilon) &= \delta(\atomize{\varphi},\epsilon)\\
\delta(\atomize{\BOX{\phi}\varphi},\epsilon) &= \true \quad \mbox{($\phi$ propositional)}\\
\delta(\atomize{\BOX{\psi?}{\varphi}},\epsilon) &= \delta(\atomize{\nnf(\lnot\psi)},\epsilon) \lor \delta(\atomize{\varphi},\epsilon)\\
\delta(\atomize{\BOX{\rho_1+\rho_2}{\varphi}},\epsilon) &= 
\delta(\atomize{\BOX{\rho_1}\varphi},\epsilon)\land\delta(\atomize{\BOX{\rho_2}\varphi},\epsilon)\\
\delta(\atomize{\BOX{\rho_1;\rho_2}{\varphi}},\epsilon) &= 
\delta(\atomize{\BOX{\rho_1}\BOX{\rho_2}\varphi},\epsilon)\\
\delta(\atomize{\BOX{\rho^*}\varphi},\epsilon) &= \delta(\atomize{\varphi},\epsilon)
\end{align*}
Notice also that for $\phi$ propositional, $\delta(\atomize{\phi},\Pi) = \delta(\atomize{\DIAM{\phi}\ttrue},\Pi)$ and $\delta(\atomize{\phi},\epsilon) = \delta(\atomize{\DIAM{\phi}\ttrue},\epsilon)$, as a consequence of the equivalence $\phi\equiv\DIAM{\phi}\ttrue$.

\newcommand{\algoname}{\textsc{{\LDLf}2\NFA}}
\renewcommand{\algorithmicrequire}{\textbf{Input:}}
\renewcommand{\algorithmicensure}{\textbf{Output:}}
\algrenewcommand\algorithmicindent{1em}
 \begin{figure}[!t]
\begin{algorithmic}[1]
  \small \State\textbf{algorithm} \algoname() \\\textbf{input} \LTLf
  formula $\varphi$ \\\textbf{output} \NFA $A_\varphi =
  (2^\Prop,\S,\{s_0\},\varrho,\{s_f\})$ \State $s_0 \gets
  \{\atomize{\varphi}\}$ \Comment{single initial state} \State $s_f
  \gets \emptyset$ \Comment{single final state} \State $\S \gets
  \{s_0,s_f\}$, $\varrho \gets \emptyset$ \While{($\S$ or $\varrho$
    change)} 
  \If{($q\in \S$ and $q'\models \bigwedge_{(\atomize{\psi}\in q)}
    \delta(\atomize{\psi},\Theta)$)}

  \State $\S \gets \S \cup \{q'\}$ \Comment{update set of states}
  \State $\varrho \gets \varrho \cup \{ (q,\Theta,q')\}$ \Comment{update
    transition relation}
  \EndIf
  \EndWhile
\end{algorithmic}
 \vspace{-.3cm}
 \caption{\NFA construction }\label{fig:algo}
 \end{figure}
\noindent Using the auxiliary function $\delta$ we can build the \NFA
$A_\varphi$ of an \LDLf formula $\varphi$ in a forward fashion as
described in Figure~\ref{fig:algo}), where:
states of $A_\varphi$ are sets of atoms (recall that each atom is
quoted $\varphi$ subformulas) to be interpreted as a conjunction; the
empty conjunction $\emptyset$ stands for $\true$; $\Theta$ is either a propositional interpretation $\Pi$ over $\Prop$ or the empty trace $\epsilon$ (this gives rise to epsilon transition either to true or false) and $q'$ is a set of quoted subformulas of $\varphi$ that denotes a
minimal interpretation such that $q'\models
\bigwedge_{(\atomize{\psi}\in q)} \delta(\atomize{\psi},\Theta)$. (Note:
we do not need to get all $q$ such that $q'\models
\bigwedge_{(\atomize{\psi}\in q)} \delta(\atomize{\psi},\Theta)$, but
only the minimal ones.)  Notice that trivially we have
$(\emptyset,a,\emptyset)\in\varrho$ for every $a\in\Sigma$.

The algorithm \algoname\ terminates in at most exponential number of
steps, and generates a set of states $\S$ whose size is at most
exponential in the size of  $\varphi$.

\begin{theorem}
  Let $\varphi$ be an \LDLf formula and $A_\varphi$ the \NFA
  constructed as above. Then $\pi\models\varphi \mbox{ iff } \pi\in
  L(A_\varphi)$ for every finite trace $\pi$.
\end{theorem}
\begin{proofsk}
  Given a \LDLf formula $\varphi$, $\delta$ grounded on the
  subformulas of $\varphi$ becomes the transition function of the
  \AFW, with initial state $\atomize{\varphi}$ and no final states,
  corresponding to $\varphi$ \cite{DegVa13}. Then \algoname\
  essentially transforms the \AFW into a \NFA.
\end{proofsk}


Notice that above we have assumed to have a special proposition
$\Last\in\Prop$.  If we want to remove such an assumption, we can
easily transform the obtained automaton $A_\varphi =
(2^\Prop,\S,\{\atomize{\varphi}\},\varrho,\{\emptyset\})$ into the new
automaton
\[A'_\varphi =
(2^{\Prop-\{\Last\}},\S\cup\{\textit{ended}\},\{\atomize{\varphi}\},\varrho',\{\emptyset,
\textit{ended}\})\] where:
$(q, \Pi', q') \in \varrho'$ iff $(q, \Pi', q')\in\varrho$, or $(q,
\Pi'\cup\{\Last\}, \true) \in \varrho$ and $q'=\textit{ended}$.

It is easy to see that the \NFA obtained can be built on-the-fly while
checking for nonemptiness, hence we have:
\begin{theorem}
  Satisfiability of an \LDLf formula can be checked in PSPACE by
  nonemptiness of $A_\varphi$ (or $A'_\varphi$).
\end{theorem}
Considering that it is known that satisfiability in \LDLf is a
PSPACE-complete problem we can conclude that the proposed construction
is optimal wrt computational complexity for satisfiability, as well as
for validity and logical implication which are linearly reducible to
satisfiability in \LDLf (see~\cite{DegVa13} for details).

\section{Run-time Monitoring}
\label{sec:rtm}
From an high-level perspective, the monitoring problem amounts to
observe an evolving system execution and report the violation or
satisfaction of properties of interest at the earliest possible
time. As the system progresses, its execution trace increases, and at
each step the monitor checks whether the trace seen so far conforms to
the properties, by considering that the execution can still continue.
This evolving aspect has a significant impact on the monitoring output:
at each step, indeed, the outcome may have a
degree of uncertainty due to the fact that future executions are yet
unknown.

Several variant of monitoring semantics have been proposed (see
\cite{Bauer2010:LTL} for a survey). In this paper we adopt the
semantics in \cite{MMW11}, which is basically the finite-trace variant
of the RV semantics in \cite{Bauer2010:LTL}: Given a \LTLf or
\LDLf formula $\varphi$, each time the system evolves, the monitor
returns one among the following truth values:
\begin{compactitem}
\item $ [\varphi]_{RV}=\temptrue$, meaning that the current execution trace \emph{temporarily satisfies} $\varphi$, i.e.,
  it is currently compliant with $\varphi$, but there is a
  possible system future prosecution which may lead to falsify $\varphi$;

\item $[\varphi]_{RV}=\tempfalse$, meaning that the current trace  \emph{temporarily falsify} $\varphi$, i.e.,
  $\varphi$ is not current compliant with $\varphi$, but there is a
  possible system future prosecution which may lead to satisfy $\varphi$;

\item $[\varphi]_{RV}=\true$, meaning that the current trace  \emph{satisfies} $\varphi$ and it will always do,
  no matter how it proceeds;

\item $[\varphi]_{RV}=\false$, meaning that the current trace  \emph{falsifies} $\varphi$ and it will always do,
  no matter how it proceeds.
\end{compactitem}
The first two conditions are unstable because they may change into any
other value as the system progresses. This reflects the general
unpredictability of system possible executions. Conversely, the other
two truth values are stable since, once outputted, they will not
change anymore. Observe that a stable truth value can be reached in
two different situations:
 \begin{inparaenum}[\it (i)] 
 \item when the system execution terminates;
\item when the formula that is being monitored can be fully evaluated
  by observing a partial trace only.
\end{inparaenum} 
The first case is indeed trivial, as when the
execution ends, there are no possible future evolutions and hence it
is enough to evaluate the finite (and now complete) trace seen so far
according to the  \LDLf semantics. In the
second case, instead, it is irrelevant whether the systems continues
its execution or not, since some \LDLf properties, such as
eventualities or safety properties, can be fully evaluated as soon as
something happens, e.g., when the eventuality is verified or the
safety requirement is violated.
Notice also that when a stable value is outputted, the monitoring
analysis can be stopped.

From a more theoretical viewpoint, given an \LDLf property $\varphi$, the
monitor looks at the trace seen so far, assesses if it is a \emph{prefix} of a complete trace 
not yet completed, and categorizes it according to its
potential for satisfying or violating $\varphi$ in the future. We call a prefix \emph{possibly good} for an \LDLf formula
$\varphi$ if there exists an extension of it which satisfies $\varphi$.
More precisely,
  given an \LDLf formula $\varphi$, we define the set of \emph{possibly
    good prefixes for $\L(\varphi)$} as the set 
\begin{equation}\label{def:possGood}
\L_{\possgood}(\varphi) = \{\pi
  \mid \exists \pi'. \pi \pi' \in \L(\varphi)\}
\end{equation}
Prefixes for which every possible extension satisfies $\varphi$ are instead
called \emph{necessarily good}. More precisely,
  given an \LDLf formula $\varphi$, we define the set of
  \emph{necessarily good prefixes for $\L(\varphi)$} as the set
\begin{equation}\label{def:necGood}
  \L_{\necgood}(\varphi) = \{\pi\mid \forall \pi'. \pi \pi' \in \L(\varphi)\}.
\end{equation}

The set of \emph{necessarily bad prefixes} $\L_\necbad(\varphi)$ can be
defined analogously as
\begin{equation}\label{def:necBad}
  \L_{\necbad}(\varphi) = \{\pi\mid \forall \pi'. \pi \pi' \not\in \L(\varphi)\}.
\end{equation}
Observe that the necessarily bad prefixes
for $\varphi$ are the necessarily good prefixes for $\neg \varphi$, i.e., $ \L_{\necbad}(\varphi)= \L_{\necgood}(\lnot\varphi)$.


Using this language theoretic notions, we can provide a precise
characterization of the semantics four standard monitoring evaluation
functions \cite{MMW11}.
\begin{proposition}
\label{thm:RVLDL}
Let $\varphi$ be an \LDLf formula and $\pi$ a trace. Then:
\begin{compactitem}
\item $\pi \models [\varphi]_{RV}=\temptrue$ iff $\pi \in \L(\varphi) \setminus \L_\necgood(\varphi)$;
\item $\pi \models [\varphi]_{RV}=\tempfalse$ iff $\pi \in \L(\neg \varphi) \setminus \L_\necbad(\varphi)$;
\item $\pi \models [\varphi]_{RV}=\true$ iff $\pi \in \L_{\necgood}(\varphi)$;
\item $\pi \models [\varphi]_{RV}=\false$ iff $\pi \in \L_{\necbad}(\varphi)$.
\end{compactitem}
\end{proposition}
\begin{proofsk}
Immediate from the definitions in \cite{MMW11} and the language theoretic definitions above.
\end{proofsk}

We close this section by exploiting the language theoretic notions to
better understand the relationships between the various kinds of
prefixes.
We start by observing that, the set of all finite words over the alphabet $2^{\P}$ is
the union of the language of $\varphi$ and its complement
$\L(\varphi) \cup \L(\neg \varphi)=(2^{\P})^*$.
Also, any language and its complement are disjoint
$\L(\varphi) \cap \L(\neg \varphi) = \emptyset$.

Since from the definition of possibly good prefixes we have
$\L(\varphi) \subseteq \L_{\possgood}(\varphi)$ and $\L(\neg \varphi) \subseteq
\L_{\possgood}(\neg \varphi)$,  we also have that
$\L_{\possgood}(\varphi) \cup \L_{\possgood}(\neg \varphi)=(2^{\P})^*$. 
Also from the definition it is easy to see that
$\L_{\possgood}(\varphi) \cap \L_{\possgood}(\neg \varphi) =\{ \pi \mid \exists \pi'.\pi\pi' \in \L(\varphi) \land \exists
\pi''.\pi\pi'' \in \L(\neg \varphi)\}$
meaning that the set of possibly good prefixes for $\varphi$ and the set
of possibly good prefixes for $\neg \varphi$ do intersect, and in such an
intersection are paths that can be extended to satisfy $\varphi$ but can
also be extended to satisfy $\neg \varphi$. 
It is also easy to see that 
$\L(\varphi) = \L_{\possgood}(\varphi) \setminus \L(\neg \varphi).$

Turning to 
necessarily good prefixes and necessarily bad prefixes, it is easy to
see that $\L_{\necgood}(\varphi) = \L_{\possgood}(\varphi) \setminus
\L_{\possgood}(\neg \varphi)$, that $\L_{\necbad}(\varphi) =
\L_{\possgood}(\neg \varphi) \setminus \L_{\possgood}(\varphi)$, and
also that $\subseteq \L(\varphi)\;\; \text{and}\;\; \L_{\necgood}(\varphi) \not \subseteq \L(\neg
\varphi)$.

Interestingly, necessarily good, necessarily bad, possibly good prefixes partition all finite traces. Namely
\begin{proposition}
The set of all traces $(2^\P)^*$ can be partitioned into
\[\begin{array}{l@{\qquad\quad}c@{\qquad\quad}r}
\L_{\necgood}(\varphi)&
\L_{\possgood}(\varphi) \cap \L_{\possgood}(\neg \varphi)&
\L_{\necbad}(\varphi)
\end{array}
\]

\vspace*{-1cm}
\[
\text{such that }
\begin{array}[t]{l}
\L_{\necgood}(\varphi) \cup (\L_{\possgood}(\varphi) \cap \L_{\possgood}(\neg
  \varphi)) \cup \L_{\necbad}(\varphi) = (2^\P)^*\\
\L_{\necgood}(\varphi) \cap (\L_{\possgood}(\varphi) \cap \L_{\possgood}(\neg \varphi)) \cap \L_{\necbad}(\varphi) =
\emptyset.
\end{array}
\]
\end{proposition}
\begin{proofsk}
Follows from the definitions of the necessarily good, necessarily bad, possibly good prefixes of $\L(\varphi)$ and  $\L(\lnot\varphi)$.
\end{proofsk}

\section{Runtime Monitors in \LDLf}
\label{sec:monitor}
As discussed in the previous section the core issue in monitoring is
prefix recognition. \LTLf is not expressive enough to talk about
prefixes of its own formulas. Roughly speaking, given a \LTLf formula,
the language of its possibly good prefixes for cannot be in
general described as an \LTLf formula. For such a reason, building a
monitor usually requires direct manipulation of the automaton for
$\phi$.

\LDLf instead can capture any nondeterministic automata as a formula, and it has 
 the
capability of expressing properties on prefixes. We 
can exploit such an extra expressivity to capture
the monitoring condition in a direct and elegant way.
We start by showing how to construct formulas representing (the
language of) prefixes of other formulas, and then we prove how
use them for the monitoring problem.

More precisely, given an \LDLf formula $\varphi$, it is possible to
express the language $\L_{possgood}(\varphi)$ with an \LDLf formula
$\varphi'$. Such a formula is obtained in two steps.

\begin{lemma}
  Given a \LDLf formula $\varphi$, there exists a regular expression
  $\pref_{\varphi}$ such that $\L(\pref_{\varphi}) =
  \L_{\possgood}(\varphi)$.
\end{lemma}
\begin{proofsk}
  The proof is constructive. We can build the \NFA $\A$ for $\varphi$
  following the procedure in \cite{DegVa13}. We then set as final all
  states of $\A$ from which there exists a non-zero length path to a
  final state. This new finite state machine $\A_{\possgood}(\varphi)$
  is such that
  $\L(\A_{\possgood}(\varphi))=\L_{\possgood}(\varphi)$. Since \NFA
  are exactly as expressive as regular expressions, we can translate
  $\A_{\possgood}(\varphi)$ to a regular expression $\pref_{\varphi}$.
\end{proofsk}

Given that \LDLf is as expressive as regular expression
(cf.~\cite{DegVa13}), we can translate $\pref_{\varphi}$ into an
equivalent \LDLf formula, as the following states.

\begin{theorem}
  \label{th:prefFormula}
  Given a \LDLf formula $\varphi$,
  \[\begin{array}{l}
    \pi\in \L_{\possgood}(\varphi)\; \text{iff}\; \pi\models \DIAM{\pref_{\varphi}}\Endt\\
    \pi\in \L_{\necgood}(\varphi)\; \text{iff}\; \pi\models \DIAM{\pref_{\varphi}}\Endt\land \lnot \DIAM{\pref_{\lnot\varphi}}\Endt\\
  \end{array}
  \]

\end{theorem}
\begin{proofsk}
  Any regular expression $\rho$, and hence any regular language, can be captured in \LDLf as $\DIAM{\rho}\Endt$.
Hence the language $\L_{\necgood}(\varphi)$ can be captured by $\DIAM{\pref_{\varphi}}\Endt$ and the language $\L_{\necgood}(\varphi)$
which is equivalent $\L_{\possgood}(\varphi) \setminus \L_{\possgood}(\neg \varphi)$ can be captured by captured by $\DIAM{\pref_{\varphi}}\Endt\land \lnot \DIAM{\pref_{\lnot\varphi}}\Endt$.
\end{proofsk}

In other words, given a \LDLf formula $\varphi$, formula
$\varphi'=\DIAM{\pref_{\varphi}}\Endt$ is a \LDLf formula such that
$\L(\varphi') = \L_{possgood}(\varphi)$. Similarly for $\L_{\necgood}(\varphi)$.

Exploiting this result, and the results in
Proposition \ref{thm:RVLDL}, we reduce RV monitoring to the
standard evaluation of \LDLf formulas over a (partial) trace. Formally: 

\begin{theorem}
\label{thm:rv-ltl}
Let $\pi$ be a (typically partial) trace. The following equivalences hold:
  \begin{compactitem}
  \item $\pi \models [\varphi]_{RV}=\temptrue$\; iff\; $\pi \models
    \varphi \land \DIAM{\pref_{\lnot\varphi}}\Endt$;
  \item $\pi \models [\varphi]_{RV}=\tempfalse$\; iff\; $\pi \models
    \lnot\varphi \land \DIAM{\pref_{\varphi}}\Endt$;
  \item $\pi \models [\varphi]_{RV}=\true$\; iff\;
    $\DIAM{\pref_{\varphi}}\Endt\land \lnot
    \DIAM{\pref_{\lnot\varphi}}\Endt$;
  \item $\pi \models [\varphi]_{RV}=\false$\; iff\;
    $\DIAM{\pref_{\lnot\varphi}}\Endt\land \lnot
    \DIAM{\pref_{\varphi}}\Endt$.
  \end{compactitem}
\end{theorem}
\begin{proofsk}
  Follows from Proposition~\ref{thm:RVLDL} and
  Theorem~\ref{th:prefFormula} using the language theoretic
  equivalences discussed in Secton~\ref{sec:rtm}.
\end{proofsk}


\renewcommand{\tabularxcolumn}[1]{>{\arraybackslash}m{#1}}

\newcommand{\dm}{\M}
\newcommand{\supp}[2]{\textsc{supp}_{#1}(#2)}
\newcommand{\ftt}[1]{\textsc{to-ft}(#1)}
\newcommand{\itt}[1]{\overline{#1}}
\newcommand{\ft}[1]{\textsc{to-\LTLf}(#1)}
\newcommand{\unique}[1]{\xi_{#1}}
\newcommand{\modelsi}{\models_{\text{LTL}}}

\section{Monitoring Declare Constraints and Metaconstraints}
\label{sec:monitoringDeclare}
We now ground our monitoring approach to the case of \declare monitoring.
\declare\footnote{\url{http://www.win.tue.nl/declare/}} is a language and framework for the declarative,
constraint-based modelling of 
processes and
services. 
A thorough
treatment of constraint-based processes can be found in \cite{Pes08,Mon10}. 
As a modelling language, \declare takes a complementary approach
 to that of classical, imperative process modeling, in which all allowed control-flows
among tasks must be explicitly
represented, and every other execution trace is implicitly considered as
forbidden. 
Instead of this procedural and ``closed'' approach,
\declare has a declarative, ``open'' flavor: the agents responsible
for the process execution can freely choose how to perform the
involved tasks, provided that the resulting execution trace complies
with the modeled business constraints.
  This is the reason why, alongside
 traditional control-flow constraints such as sequence (called in
 \declare \emph{chain succession}), \declare supports a plethora of peculiar
 constraints that do not impose specific temporal orderings, or that
 explicitly account with negative information, i.e., prohibition of
 task execution.

Given a set $\Prop$ of tasks, a \declare model is a set $\C$ of \LTLf (and hence \LDLf) constraints over $\Prop$, used to restrict the
  allowed execution traces. 
Among all possible \LTLf constraints, some specific \emph{patterns} have been
singled out as particularly meaningful for expressing \declare
processes, taking inspiration from \cite{DwAC99}. Such patterns are grouped into four
families:
\begin{inparaenum}[\it (i)]
\item \emph{existence} (unary) constraints, stating that the target
  task must/cannot be executed (a certain amount of times);
\item \emph{choice} (binary) constraints, modeling choice of
  execution;
\item \emph{relation} (binary) constraints, modeling that whenever the
  source task is executed, then the target task must also be executed
  (possibly with additional requirements);
\item \emph{negation} (binary) constraints, modeling that whenever the
  source task is executed, then the target task is prohibited
  (possibly with additional restrictions).
\end{inparaenum}
Table~\ref{tab:constraints} reports some of these
patterns. 

\begin{example}
\label{ex:declare}
Consider a fragment of a purchase order process, where we consider three key
business constraints. First, an order can be closed at most once. In
\declare, this can be tackled with a \constraint{absence 2} constraint, 
visually and formally represented as:
\vspace*{-.2cm}
\[
\existenceformula{\absenceN{1}}{close order} \qquad 
\varphi_{close} = \neg \Diamond (\activity{close order} \land \Next
\Diamond \activity{close order} )
\]
Second, an order can be canceled only until it
is closed. This can be captured by a \constraint{negation succession}
constraint, which states that after the order is closed, it cannot be
canceled anymore:
\[
\negationformula{\succession}{close order}{cancel order} \qquad
\varphi_{canc} = \Box (\activity{close order} \limp \neg
\Diamond \activity{cancel order} )
\]
Finally, after the order is closed, it becomes possible to do
supplementary payments, for various reasons (e.g., to speed up the
delivery of the order). 
\[
\relationformula{\precedence}{close order}{pay suppl} ~
\varphi_{pay} = (\neg \activity{pay suppl}  \Until \activity{close
  order}) \lor \neg \Diamond \activity{close order}
\]
\end{example}

Beside modeling and enactment of constraint-based processes, previous
works have also focused on runtime verification of \declare models. A
family of \declare monitoring approaches rely on the original \LTLf
formalization of \declare, and employ corresponding automata-based
techniques to track running process instances and check whether they
satisfy the modeled constraints or not \cite{MMW11,MWM12}. Such techniques have
been in particular used for:
\begin{compactitem}
\item Monitoring single \declare constraints so as to provide a fine-grained
  feedback; this is done by adopting the RV semantics for \LTLf,
  and tracking the evolution each constraint through the four RV
  truth values.
\item Monitoring the global \declare model by considering all
  its constraints together (i.e., constructing a \DFA for the
  conjunction of all constraints); this is important for computing the \emph{early
    detection} of violations, i.e.,  violations that cannot be
  explicitly found in the execution trace collected so far, but that 
  cannot be avoided in the future. 
\end{compactitem}

We now discuss how \LDLf can be adopted for monitoring \declare
constraints, with a twofold advantage. First, as shown in
Section~\ref{sec:monitor}, \LDLf is able to encode the RV
semantics directly into the logic, without the need of introducing
ad-hoc modifications in the corresponding standard logical services.
 Second, beside being able to reconstruct
all the aforementioned monitoring techniques, our approach also provides a
declarative, well-founded basis for monitoring metaconstraints, i.e.,
constraints that involve both the execution of tasks and the monitoring
outcome obtained by checking other constraints.  

\paragraph{\bf Monitoring Declare Constraints with \LDLf.}
Since \LDLf includes \LTLf, \declare constraints can be directly encoded in
\LDLf using their standard formalization \cite{PesV06,MPVC10}. Thanks
to the translation into {\NFA}s discussed in
Section~\ref{sec:automaton} (and, if needed, their determinization into
corresponding {\DFA}s), the obtained automaton can then be used to
check whether a (partial) finite trace satisfies this constraint or
not. This is not very effective, as the approach does not support
the detection of fine-grained truth values as those of RV. 
By relying on Theorem~\ref{thm:rv-ltl}, however, we can reuse the same
technique, this time supporting all RV. In fact, by formalizing
the good prefixes of each \declare pattern, we can immediately
construct the four \LDLf formulas that embed the different RV
truth values, and check the current trace over each of the
corresponding automata. Table~\ref{tab:constraints} reports the good
prefix characterization of some of the \declare patterns; it can be
seamlessly extended to all other patterns as well.  

\begin{example}
Let us consider the \constraint{absence 2} constraint $\varphi_{close}$ in
Example~\ref{ex:declare}. Following Table~\ref{tab:constraints}, its
good prefix characterization is
$\pref_{\varphi_{close}} = \ot^*+(\ot^*;\activity{close
  order};\ot^*)$, 
where $\ot$ is a shortcut for all the tasks involved in the purchase
order process but \activity{close order}. This can be used to
construct the four formulas mentioned in Theorem~\ref{thm:rv-ltl},
which in turn provide the basis to produce, e.g., the following result:

\noindent\begin{tabularx}{\textwidth}{r|p{1.8cm}|p{2.7cm}|p{2.7cm}|p{2.7cm}}
&start &do ``close order'' & do ``pay suppl.'' & do ``close order''\\
\hline
\existenceformula{\absenceN{1}}{close order} & \multicolumn{3}{c|}{$\temptrue$} & \multicolumn{1}{c}{$\false$}\\
\hline
\end{tabularx}
\end{example}

Observe that this baseline approach can be extended along a number of
directions. For example, as shown in Table~\ref{tab:constraints}, the
majority of \declare patterns does not cover all the four RV truth
values. This is the case, e.g., for \constraint{absence 2}, which can
never be evaluated to be $\true$ (since it is always possible to
continue the execution so as to perform \activity{a} twice), nor to
$\tempfalse$ (the only way of violating the constraint is to perform
\activity{a} twice, and in this case it is not possible to ``repair''
to the violation anymore). This information can be used to restrict
the generation of the automata only to those cases that are relevant
to the constraint. Furthermore, it is possible to reconstruct exactly
the approach in \cite{MMW11}, where every state in the {\DFA}s
corresponding to the constraints to be monitored, is enriched with a
``color'' accounting for one of the four RV truth values. To do
so, we have simply to combine the four
{\DFA}s generated for each constraint. This is possible because such 
{\DFA}s are generated from formulas built on top of the good prefix
characterization of the original formula, and hence they all produce the same automaton, but with different
final states. In fact, this observation provides a formal
justification to the correctness of the approach in \cite{MMW11}.

\begin{table*}[t!]
\caption{\label{tab:constraints} Some \declare constraints, together
  with their prefix characterization, minimal bad prefix
  charaterization, and possible RV states; for each constraint, $\ot$ is
  a shortcut for ``other tasks'', i.e., tasks not involved in the constraint itself.}
\scalebox{0.7}{
\begin{tabularx}{1.43\textwidth}{@{}c@{~~}l@{~~}l@{~~} l @{~~}X@{}}
\toprule
& \textsc{name} & \textsc{notation} & $\pref$ 
& \textsc{possible RV states}\\
\midrule
\parbox[b]{4mm}{\multirow{3}{*}{\rotatebox[origin=c]{90}{\textsc{existence\!\!\!\!}}}}
&
\textbf{Existence} 
&
\existenceformula{\existenceN{1}}{a}
&
$(a+\ot)^*$
&
$\tempfalse$, $\true$
\\
\cmidrule{2-5}
&
\textbf{Absence 2} 
&
\existenceformula{\absenceN{1}}{a}
&
$\ot^*+(\ot^*;a;\ot^*)$
&
$\temptrue$, $\false$
\\
\midrule
\parbox[b]{4mm}{\multirow{2}{*}{\rotatebox[origin=c]{90}{\textsc{choice}}}}
&
\textbf{Choice} 
&
\relationformula{\choice}{a}{b} 
&
$(a+b+\ot)^*$
&
$\tempfalse$, $\true$
\\
\cmidrule{2-5}
&
\textbf{Exclusive Choice}
&
\relationformula{\exclusivechoice}{a}{b}
&
$(a+\ot)^*+(b+\ot)^*$
&
$\tempfalse$, $\temptrue$, $\false$
\\
\midrule
\parbox[b]{4mm}{\multirow{5}{*}{\rotatebox[origin=c]{90}{\textsc{relation}}}}
&
\textbf{Resp.\ existence} 
&
\relationformula{\respondedexistence}{a}{b}
&
$(a+b+\ot)^*$
&
$\temptrue$, $\tempfalse$, $\true$
\\
\cmidrule{2-5}
&
\textbf{Coexistence} 
&
\relationformula{\coexistence}{a}{b}
&
$(a+b+\ot)^*$
&
$\temptrue$, $\tempfalse$, $\true$
\\
\cmidrule{2-5}
&
\textbf{Response} 
&
\relationformula{\response}{a}{b}
&
$(a+b+\ot)^*$
&
$\temptrue$, $\tempfalse$
\\
\cmidrule{2-5}
&
\textbf{Precedence} 
&
\relationformula{\precedence}{a}{b}
&
$\ot^*;(a;(a+b+\ot)^*)^*$
&
$\temptrue$, $\true$, $\false$
\\
\cmidrule{2-5}
&
\textbf{Succession} 
&
\relationformula{\succession}{a}{b}
&
$\ot^*;(a;(a+b+\ot)^*)^*$
&
$\temptrue$, $\tempfalse$, $\false$
\\
\midrule
\parbox[b]{4mm}{\multirow{2}{*}{\rotatebox[origin=c]{90}{\textsc{negation}}}}
&
\textbf{Not Coexistence} 
&
\negationformula{\coexistence}{a}{b}
&
$(a+\ot)^*+(b+\ot)^*$
&
$\temptrue$, $\false$
\\
\cmidrule{2-5}
&
\textbf{Neg.\ Succession} 
&
\negationformula{\succession}{a}{b}
&
$(b+\ot)^*;(a+\ot)^*$
&
$\temptrue$, $\false$
\\
\bottomrule
\end{tabularx}
}
\end{table*}


\paragraph{\bf Metaconstraints.}
Thanks to the ability of \LDLf to directly encode into the logic
\declare constraints but also their RV monitoring states, we can
formalize metaconstraints that relate the RV truth values of different
constraints. Intuitively, such metaconstraints allow one to capture 
that \emph{we become interested in monitoring some constraint only when
other constraints are evaluated to be in a certain RV truth
value}. This, in turn, provides the basis to declaratively capture two
classes of properties that are of central importance in the context of
runtime verification:
\begin{compactitem}
\item \emph{Compensation constraints}, that is, constraints that should
  be enforced by the agents executing the process in the case other
  constraints are violated, i.e., are evaluated to be $\false$. Previous works have been tackled this issue
 through ad-hoc techniques, with no declarative counterpart \cite{MMW11,MWM12}.
\item Recovery mechanisms resembling \emph{contrary-to-duty
    obligations} in legal reasoning \cite{PrS96}, i.e., obligations
  that are put in place only when other obligations are not met.
\end{compactitem}
Technically, a generic form for metaconstraints is the pattern $\Phi_{pre} \limp \Psi_{exp}$, 
where:
\begin{compactitem}
\item $\Phi_{pre}$ is a boolean formula, whose atoms are membership
  assertions of the involved constraints to the RV truth values;
\item $\Psi_{exp}$ is a boolean formula whose atoms are the
  constraints to be enforced when $\Phi_{pre}$ evaluates to true.
\end{compactitem}
 This pattern can be used, for example, to state that whenever constraints
$c_1$ and $c_2$ are permanently violated, then either constraint $c_3$
or $c_4$ have to be enforced. Observe that the metaconstraint so
constructed is a standard \LDLf formula. Hence, we can reapply 
Theorem~\ref{thm:rv-ltl} to it, getting four \LDLf formulas that can
be used to track the evolution of the metaconstraint among the four RV values.
\begin{example}
\label{ex:comp}
Consider the \declare constraints of Example \ref{ex:declare}.  We
want to enhance it with a compensation constraint stating that
whenever $\varphi_{canc}$ is violated (i.e., the order is canceled
after it has been closed), then a supplement payment must be
issued. This can be easily captured in \LDLf as follows. First of all,
we model the compensation constraint, which corresponds, in this case,
to a standard \constraint{existence} constraint over the \activity{pay
supplement} task. Let $\varphi_{dopay}$ denote the \LTLf formalization
of such a compensation constraint. Second, we capture the intended
compensation behavior by using the following \LDLf metaconstraint:
\[
\{[\varphi_{canc}]_{RV} = \false\}  \limp \varphi_{dopay}
\]
which, leveraging Theorem~\ref{thm:rv-ltl}, corresponds to the
standard \LDLf formula:
\[
\footnotesize
(\DIAM{\pref_{\lnot\varphi_{canc}}}\Endt\land \lnot
    \DIAM{\pref_{\varphi_{canc}}}\Endt) \limp \varphi_{dopay}
\]
\end{example}
A limitation of this form of metaconstraint is that the right-hand
part $\Psi_{exp}$ is monitored \emph{from the beginning of the
  trace}. This is acceptable in many cases. E.g., in
Example~\ref{ex:comp}, it is ok if the user already paid a
supplement before the order cancelation caused constraint
$\varphi_{canc}$ to be violated. In other situations, however, this is
not satisfactory, because we would like to enforce the compensating
behavior only \emph{after} $\Phi_{pre}$ evaluates to true, e.g., after
the violation of a given constraint has been detected. In general, we
can extend the aforementioned metaconstraint pattern as follows:
$\Phi_{pre} \limp \BOX{\rho} \Psi_{exp}$, where $\rho$ is a regular
expression denoting the paths after which $\Psi_{exp}$ is expected to
be enforced. 

By constructing $\rho$ as the regular expression
accounting for the paths that make $\Phi_{pre}$ true, we can then
exploit this improved metaconstraint to  express that $\Psi_{exp}$
is expected to become true after all prefixes of the current trace
that made $\Phi_{pre}$ true.

\begin{example}
We modify the compensation constraint of Example~\ref{ex:comp}, so as
to reflect that when a closed order is canceled (i.e.,
$\varphi_{canc}$ is violated), then a supplement must be paid
\emph{afterwards}. This is captured by the following metaconstraint:
\[
\{[\varphi_{canc}]_{RV} =\false\}  \limp
\BOX{\re_{\{[\varphi_{canc}]_{RV} = \false\}}} \varphi_{dopay}
\]
where $\re_{\{[\varphi_{canc}]=\false\}}$ denotes the regular
expression for the language $\L(\{[\varphi_{canc}]=\false\})=\L(\DIAM{\pref_{\lnot\varphi_{canc}}}\Endt\land \lnot
\DIAM{\pref_{\varphi_{canc}}}\Endt)$. This
regular expression describes all paths containing a violation for
constraint $\varphi_{canc}$.
 \end{example}

\section{Implementation}
\begin{figure}
  \centering
    \includegraphics[scale=0.5]{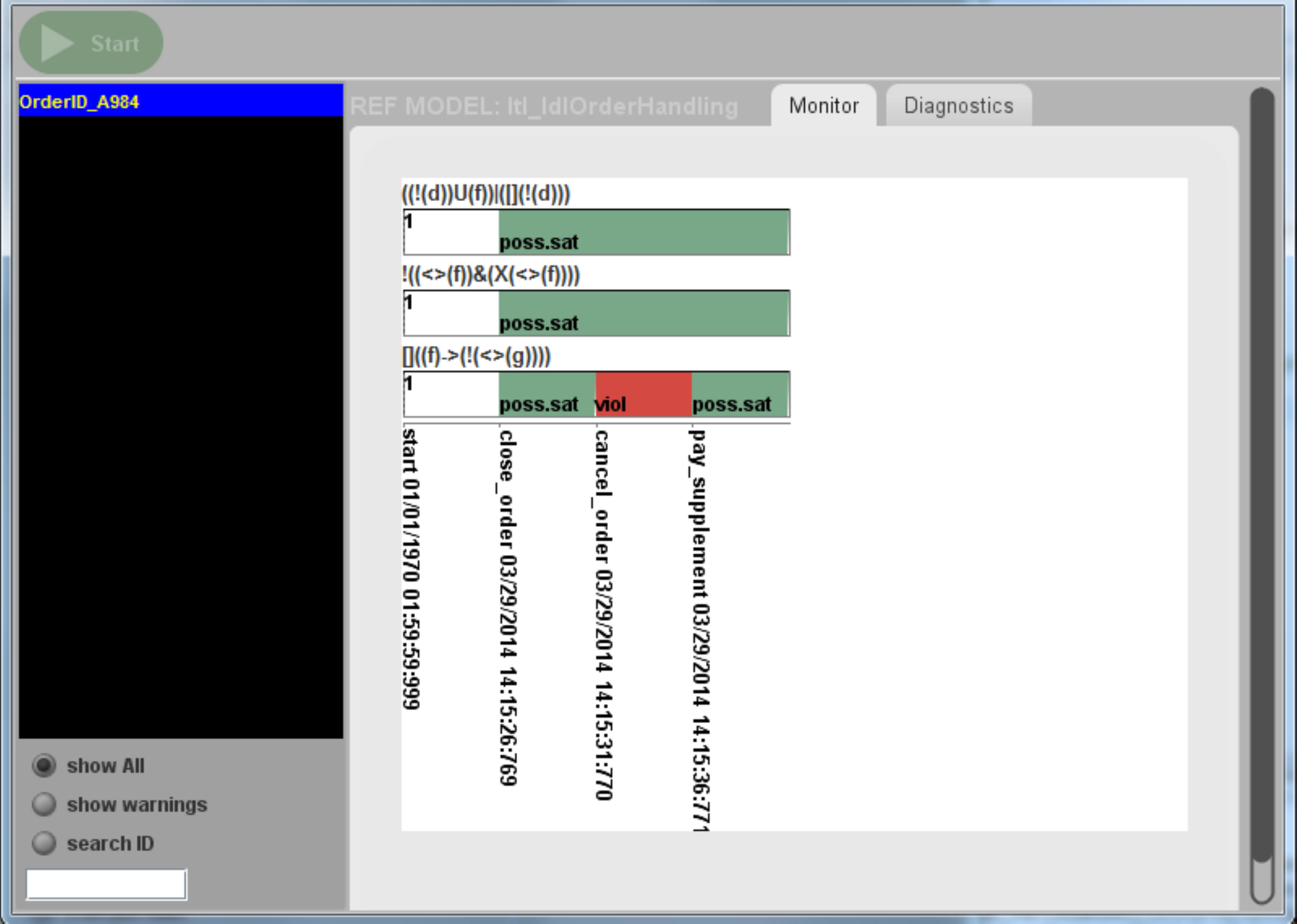}
  \caption{Screenshot of our operational support provider's output.}
\label{fig:implementation}
\end{figure}
The entire approach has been implemented as an \emph{operational decision
support (OS) provider} for the \textsc{ProM} 6 process mining
framework\footnote{\url{http://www.promtools.org/prom6/}}. \textsc{ProM}
6 provides a generic OS
environment~\cite{Westergaard2011:OS} that supports the interaction
between an external workflow management systems
at runtime (producing events) and \textsc{ProM}. In particular, it
provides an OS service that receives a stream of events
from the external world, updates and
orchestrates the registered OS providers implementing different types
of online analysis to be applied on the stream, and reports the
produced results back to the external world.

At the back-end of the plug-in, there is a software module
specifically dedicated to the construction and manipulation of {\NFA}s
from \LDLf formulas, concretely implementing the technique presented
in Section~\ref{sec:automaton}. To manipulate regular expressions and
automata, we used the fast, well-known library \texttt{dk.brics.automaton}   \cite{bricksautomaton}.

Figure~\ref{fig:implementation} shows a graphical representation of
the evolution of constraints described in Example~\ref{ex:declare} of
Section~\ref{sec:monitoringDeclare} when monitored using our operational
support provider. In the \LDLf formulas, the literals $f$, $g$ and
$d$ respectively stand for tasks \activity{close order},
\activity{cancel order}, and \activity{pay supplement}.

\section{Conclusion}

We can see the approach proposed in this paper, as an extension of the declarative process specification approach, at the basis of \declare, to \myi more powerful specification logics (\LDLf, i.e., Monadic Second Order logics over finite traces , instead of \LTLf, i.e., First Order logic logics over finite traces), \myii to monitoring constraints.
Notably, this declarative approach to  monitoring supports seamlessly monitoring metaconstraints, i.e., constraints that do not only predicate about the dynamics of task executions, but also about the truth values of other constraints. 
We have grounded this approach on \declare itself, showing how to declaratively specify compensation constraints. 


The next step will be to incorporate recovery mechanisms into the approach, in particular providing a formal underpinning to the ad-hoc recovery mechanisms studied in \cite{MMW11}. Furthermore, we intend to extend our approach to \emph{data-aware} business constraints \cite{BCDDM13,MCMM13}, mixing temporal operators with first-order queries over the data attached to the monitored events. This setting has been studied using the Event Calculus \cite{MCMM13,MMC13}, also considering some specific forms of compensation in \declare \cite{CMM08}. However, the resulting approach can only query the partial trace accumulated so far, and not reason upon its possible future continuations, as automata-based techniques are able to do. To extend the approach presented here to the case of data-aware business constraints, we will build on recent, interesting decidability results for the static verification of data-aware business processes against sophisticated variants of first-order temporal logics \cite{BCDDM13}.


\begin{small}
\paragraph{Acknowledgments.}
This research has been partially supported by the EU IP project \emph{Optique: Scalable End-user Access to
  Big Data}, grant agreement n.~FP7-318338, and by the Sapienza Award
2013 ``\emph{\textsc{spiritlets}: Spiritlet-based smart spaces}''.
\end{small}

\bibliographystyle{abbrv}

\bibliography{main-bib}

\end{document}